\documentclass[10pt,twocolumn]{article}

\usepackage[a4paper,margin=1.8cm]{geometry}
\usepackage[T1]{fontenc}
\usepackage{lmodern}
\usepackage{microtype}

\usepackage{amsmath,amssymb}
\usepackage{graphicx}
\usepackage{booktabs}
\usepackage{float}
\usepackage{stfloats}  
\usepackage{tikz}
 \usepackage{eso-pic}
\usetikzlibrary{arrows.meta, calc, positioning, shapes.geometric, fit, backgrounds}
\usepackage[font=small]{caption}

\definecolor{oiBlue}{RGB}{0,114,178}
\definecolor{oiOrange}{RGB}{230,159,0}
\definecolor{oiGreen}{RGB}{0,158,115}
\definecolor{oiVermillion}{RGB}{213,94,0}
\definecolor{panelBg}{RGB}{250,250,250}
\definecolor{panelBdr}{RGB}{218,218,218}
\definecolor{hd}{RGB}{40,40,40}
\definecolor{bd}{RGB}{60,60,60}
\definecolor{dm}{RGB}{145,145,145}
\definecolor{ag}{RGB}{115,115,115}
\definecolor{rowG}{RGB}{237,250,244}
\definecolor{rowR}{RGB}{253,241,237}
\definecolor{pillG}{RGB}{225,245,235}
\definecolor{pillGb}{RGB}{0,158,115}
\definecolor{delRed}{RGB}{220,53,69}
\definecolor{delBg}{RGB}{253,237,237}

\usepackage{enumitem}
\setlist[itemize]{topsep=2pt,itemsep=1pt,parsep=0pt,partopsep=0pt,leftmargin=*}
\setlist[enumerate]{topsep=2pt,itemsep=1pt,parsep=0pt,partopsep=0pt,leftmargin=*}

\usepackage{titlesec}
\titlespacing*{\section}{0pt}{0.9ex plus 0.2ex minus 0.2ex}{0.5ex}
\titlespacing*{\subsection}{0pt}{0.7ex plus 0.2ex minus 0.2ex}{0.35ex}

\usepackage{amsthm}
\newtheorem{invariant}{Invariant}

\usepackage[hidelinks]{hyperref}

\usepackage[numbers,sort&compress]{natbib}

\title{Separable Expert Architecture: Toward Privacy-Preserving LLM Personalization via Composable Adapters and Deletable User Proxies}
\author{
  Chris Schneider$^{1}$ \quad Philipp Schoenegger$^{1}$ \quad Ben Bariach$^{1}$ \\[1ex]
  \textnormal{$^{1}$Microsoft AI}
}
\date{}

\begin{document}
  \AddToShipoutPictureBG*{%
    \AtPageUpperLeft{%
      \raisebox{-1.4cm}{\hspace{1.4cm}%
        \begin{minipage}{5cm}%
          \includegraphics[height=0.5cm]{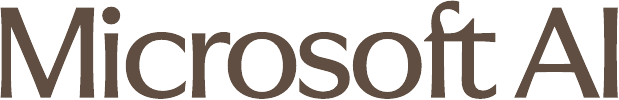}%
        \end{minipage}%
      }%
    }%
  }
\maketitle
\begin{abstract}
\textit{Current model training approaches incorporate user information directly into shared weights, making individual data removal computationally infeasible without retraining. This paper presents a three-layer architecture that decouples personal data from shared weights by combining a static base model, composable domain-expert LoRA adapters that shape behavior without imparting user data, and per-user proxy artefact whose deletion constitutes deterministic unlearning. Evaluation on Phi-3.5-mini and Llama-3.1-8B confirms per-user differentiation in which personal data influences outputs while remaining isolated, verified by a return to baseline after proxy removal (KL $\approx 0.21$ nats, $82$--$89\%$ verification pass rate) and near-zero cross-user contamination. Because user-specific information never enters shared weights, the architecture mitigates model inversion, membership inference, and training-data extraction against shared model components by construction. The approach converts machine unlearning from an intractable weight-editing problem into a deterministic deletion operation that preserves personalization alongside privacy-enhancing guarantees and is compatible with differentially private stochastic gradient descent (DP-SGD) for privacy-preserving shared model improvement.}
\end{abstract}

\begin{figure*}[b]
  \centering
  \scalebox{0.68}{%
  \begin{tikzpicture}[
    box/.style={draw, rounded corners=3pt, minimum height=0.85cm,
                line width=0.7pt, font=\small, text centered},
    sml/.style={draw, rounded corners=2pt, minimum height=0.55cm,
                line width=0.5pt, font=\scriptsize, text centered},
    arr/.style={-{Stealth[length=5pt]}, line width=0.6pt},
  ]

    \begin{scope}[on background layer]
      \fill[oiBlue!4, rounded corners=6pt]
        (-1.4,-1.7) rectangle (6.6,5.0);
      \node[font=\scriptsize\bfseries, oiBlue!60!black, anchor=north west]
        at (-1.2,4.9) {SHARED COMPONENTS};
      \node[font=\tiny, ag, anchor=north west]
        at (-1.2,4.55) {\scriptsize no user data};
    \end{scope}

    \begin{scope}[on background layer]
      \fill[oiOrange!5, rounded corners=6pt]
        (7.8,-1.7) rectangle (14.8,5.0);
      \node[font=\scriptsize\bfseries, oiOrange!70!black, anchor=north west]
        at (8.0,4.9) {USER PROXY\; $P_u$};
      \node[font=\scriptsize, ag, anchor=north west]
        at (8.0,4.55) {all user-specific data};
    \end{scope}

    \draw[delRed, line width=1.2pt, dashed, rounded corners=6pt]
      (7.7,-1.8) rectangle (14.9,5.1);
    \node[font=\scriptsize\bfseries, delRed] at (11.3,-2.1)
      {REMOVABLE --- file deletion = full erasure};

    \draw[ag, line width=0.8pt, dash pattern=on 6pt off 3pt on 2pt off 3pt]
      (7.2,-1.7) -- (7.2,5.0);

    \node[box, fill=oiBlue!12, draw=oiBlue, minimum width=5.2cm]
      (base) at (2.7,-0.8)
      {\textbf{Base Model}\; $\theta$ \color{ag}\small\;(shared)};

    \node[box, fill=oiGreen!8, draw=oiGreen!70!black, minimum width=3.8cm]
      (router) at (2.7,3.8)
      {\textbf{Expert Router}};

    \node[sml, fill=oiGreen!12, draw=oiGreen, minimum width=2.2cm]
      (e1) at (1.0,2.6) {$E_1$\;Security};
    \node[sml, fill=oiGreen!12, draw=oiGreen, minimum width=2.2cm]
      (e2) at (4.4,2.6) {$E_2$\;Code};
    \node[sml, fill=oiGreen!12, draw=oiGreen, minimum width=2.2cm]
      (e3) at (1.0,1.6) {$E_3$\;Data};
    \node[sml, fill=oiGreen!12, draw=oiGreen, minimum width=2.2cm]
      (e4) at (4.4,1.6) {$E_4$\;General};

    \draw[arr, oiGreen!70!black] (router.south) -- ++(0,-0.15) -| (e1.north);
    \draw[arr, oiGreen!70!black] (router.south) -- ++(0,-0.15) -| (e2.north);

    \begin{scope}[on background layer]
      \draw[oiGreen!40, rounded corners=4pt, line width=0.5pt]
        ([shift={(-0.2,0.15)}]e1.north west)
        rectangle
        ([shift={(0.2,-0.15)}]e4.south east);
    \end{scope}

    \node[sml, fill=white, draw=oiGreen!60!black, minimum width=3.0cm]
      (wmrg) at (2.7,0.65) {$W_{\text{base}} + \sum w_i B_i A_i$};
    \draw[arr, oiGreen!60!black] (e3.south) |- (wmrg.west);
    \draw[arr, oiGreen!60!black] (e4.south) |- (wmrg.east);
    \draw[arr, oiBlue!70!black] (wmrg) -- (base);

    \node[sml, fill=oiOrange!15, draw=oiOrange, minimum width=4.8cm,
          minimum height=0.7cm]
      (rbias) at (11.3,3.7)
      {\textbf{Routing bias}\; $\mathbf{b}_u \in \mathbb{R}^k$};
    \node[font=\normalsize, ag]
      at (11.3,3.1)
      {domain preference scores};

    \node[sml, fill=oiOrange!15, draw=oiOrange, minimum width=4.8cm,
          minimum height=0.7cm]
      (plora) at (11.3,0.7)
      {\textbf{Personal LoRA}\; $L_u = (B_u, A_u)$};
    \node[font=\normalsize, ag]
      at (11.3,0.0)
      {user-specific weight residuals};

    \node[sml, fill=oiOrange!15, draw=oiOrange, minimum width=4.8cm,
          minimum height=0.7cm]
      (svec) at (11.3,-0.8)
      {\textbf{Steering vectors}\; $\{s_u^\ell\}_{\ell \in \mathcal{L}}$};
    \node[font=\normalsize, ag]
      at (11.3,-1.5)
      {style and preference modifiers};

    \draw[arr, oiOrange!80!black] (rbias.west) -- (rbias.west -| router.east);

    \draw[arr, oiOrange!80!black]
      (plora.west) |- (wmrg.east);
    \path let \p1 = (plora.west) in
      node[
        font=\scriptsize,
        oiOrange!80!black,
        anchor=south,
        yshift=2pt
      ]
      at (7.2,\y1) {merge};

    \draw[arr, oiOrange!80!black]
      (svec.west)
      -- node[midway, above, font=\scriptsize, fill=white, inner sep=1.5pt] {inject}
         (svec.west -| base.east);

    \node[font=\small\bfseries, hd, yshift=-3pt] (qin) at (-2.8,3.9) {Query $q$};
    \draw[arr, hd] (qin) -- (router.west);
    \node[font=\small\bfseries, hd] (out) at (-2.8,-0.8) {Output};
    \draw[arr, hd] (base.west) -- (out);

  \end{tikzpicture}%
  }
  \caption{Separable Expert Architecture. Shared components (left) contain no user-specific information: a frozen base model, four domain-expert LoRA adapters selected by a per-query router, and a weighted merge. The per-user proxy (right, dashed red border) holds three deletable personalization mechanisms (routing bias, personal LoRA, and contrastive steering vectors) that compose with shared components at inference via cross-boundary arrows. The vertical dashed line marks the separation boundary, where deleting the proxy directory removes all user-specific influence with zero retraining.}
  \label{fig:architecture}
\end{figure*}
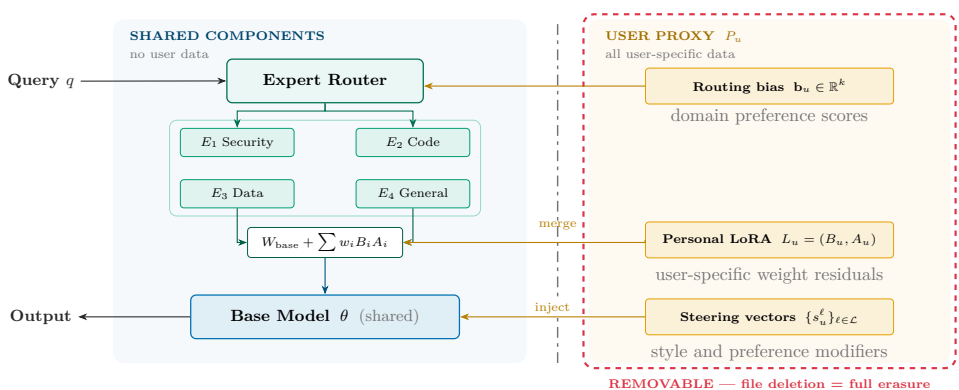

\section{Introduction}

As LLM personalization becomes widely used, a growing body of work has demonstrated that user preferences can be captured through retrieval-augmented profiles \citep{salemi2024lamp}, post-hoc parameter merging \citep{jang2023personalized}, and personalized reward learning \citep{li2024prlhf,poddar2024vpl}. While some of these approaches operate at the prompt level (e.g., retrieval-augmented profiles), many encode user-specific information into model weights $\theta$ via fine-tuning, producing models whose parameters entangle contributions from many users. When a user later requests deletion it is unclear how one can remove their data from a model whose weights have been shaped by thousands of users simultaneously. 

This suggests that there is a fundamental tension between personalization and data deletion in the context of modern LLMs. When user preferences are distributed across shared weights, deletion requires identifying and removing each user's contribution, a problem that has shown to be computationally intractable without full retraining \citep{bourtoule2021machine}. Exact unlearning methods like SISA \citep{bourtoule2021machine} require maintaining independently trained model shards, while approximate methods offer no formal removal guarantees \citep{golatkar2020eternal}. LLM-specific approaches face additional difficulties: Gradient ascent can cause catastrophic collapse in certain unlearning configurations \citep{zhang2024npo}, and representation-level methods like RMU \citep{li2024wmdp} still modify shared weights. This problem is compounded by extraction attacks, including model inversion \citep{fredrikson2015model}, training data extraction \citep{carlini2021extracting,nasr2023scalable}, and membership inference \citep{shokri2017membership}, which can recover private information from weight-encoded personalization,making it a privacy issue even absent deletion requests. To illustrate this, consider a personalized assistant that has learned a user's medical vocabulary preferences through fine-tuning. Even after the user requests deletion, membership inference attacks could reveal whether that user's data was part of the training set, while training data extraction could recover specific preference examples, all because the user's influence remains distributed across millions of shared parameters.

In order to address this issue, we propose the Separable Expert Architecture (SEA), a design that aims to satisfy both \emph{personalization} and \emph{deletability} simultaneously. The core contribution is that if user-specific information never enters shared weights, ``unlearning'' is essentially just deletion. Rather than trying to surgically undo weight entanglement after the fact, this approach prevents entanglement from occurring in the first place. In other words, this requires an architecture where personalization is \emph{compositional}, i.e., assembled at inference time from separable, deletable components, rather than \emph{absorptive}, where preferences are baked into shared parameters.

\textbf{Contributions.} We make three contributions:
\begin{enumerate}
  \item A \textbf{three-layer composition architecture} where a base model (frozen, shared) is augmented by domain-expert LoRA adapters (shared, dynamically weighted by a query router) and per-user \emph{proxy artifacts}, which are isolated directories containing a routing bias vector, contrastive steering vectors, and a personal LoRA adapter (${\sim}$2--5~MB per user in our configuration). The architecture maintains a strict invariant: All user-specific information resides in a deletable artifact that never enters shared weights (\S\ref{sec:architecture}).
  \item A \textbf{deletion protocol} that reduces user removal to filesystem deletion of the proxy directory followed by noise-calibrated KL-divergence verification against a non-personalized baseline, requiring no retraining (\S\ref{sec:deletion}) at all.
  \item Additional \textbf{empirical evidence} across Phi-3.5-mini and Llama-3.1-8B with four domain experts and four synthetic user profiles, demonstrating measurable personalization, verified deletion (82--89\% verification pass rate), and clean cross-user isolation (contamination $\leq 0.05$ in point estimates) (\S\ref{sec:results}).
\end{enumerate}

\textbf{Related Work.} Research on machine unlearning has shown that surgical removal of user influence from model weights is fundamentally hard, whether through exact retraining \citep{bourtoule2021machine} or efficient approximate deletion \citep{ginart2019making}, approximate gradient manipulation \citep{golatkar2020eternal,graves2021amnesiac}, LLM-specific methods such as model-generated knowledge replacement \citep{eldan2023harrypotter}, NPO \citep{zhang2024npo}, or representation-level unlearning \citep{li2024wmdp}. On the other hand, the infrastructure for composable adapter stacks has matured substantially: LoRA \citep{hu2022lora} and QLoRA \citep{dettmers2023qlora} enable efficient adapter training, LoraHub \citep{huang2024lorahub} and task arithmetic \citep{zhang2023composing,ilharco2022editing} demonstrate multi-adapter composition, and S-LoRA \citep{sheng2024slora} enables serving thousands of concurrent adapters from a single base model while Punica \citep{chen2023punica} provides efficient multi-tenant batching via segmented gather-matrix-vector kernels. Activation steering methods, including Contrastive Activation Addition \citep{panickssery2024steering} and Inference-Time Intervention \citep{li2023iti}, show that behavioral modification without weight changes can be both effective and relatively lightweight. LLM personalization approaches, including LaMP \citep{salemi2024lamp}, Personalized Soups \citep{jang2023personalized}, P-RLHF \citep{li2024prlhf}, and VPL \citep{poddar2024vpl}, capture user preferences through various mechanisms. However, none of these approaches architecturally separates user state from shared weights, meaning that deletion would require either retraining or approximate weight modification, the same intractable operations the unlearning literature has already identified as problematic \citep{bourtoule2021machine,golatkar2020eternal}. Adding a deletion mechanism post hoc does not resolve this as the entanglement occurs during training, and no inference-time wrapper can undo it. The infrastructure for composable, per-user adapter stacks exists, but what is largely missing is a \emph{deletion-aware} composition design that prevents entanglement from occurring in the first place. SEA bridges this gap by ensuring that personalization state is architecturally separable from shared model components.

In the rest of the paper, we go through the architecture and deletion protocol of the SEA (\S\ref{sec:architecture}), the experimental setup (\S\ref{sec:setup}), and the results (\S\ref{sec:results}), before closing with a discussion of implications and limitations (\S\ref{sec:discussion}).

\section{Architecture}
\label{sec:architecture}

In this section, we present SEA's three-layer composition architecture and its core design invariant. The central claim is that the user-specific information has to be structurally separated from shared model components such that deletion becomes a deterministic filesystem operation rather than an approximate weight-modification procedure. We first state the invariant (\S\ref{sec:invariant}), then describe the three composition layers (\S\ref{sec:layers}), detail the inference pipeline (\S\ref{sec:pipeline}), and lastly present the deletion protocol (\S\ref{sec:deletion}).

\subsection{Design Invariant}
\label{sec:invariant}

SEA maintains a strict architectural invariant that distinguishes it from approximate unlearning approaches and provides the basis for the deletion protocol:

\begin{invariant}[Separation]
\label{inv:separation}
All user-specific information resides in an isolated, deletable proxy artifact. Shared model components (the base model and expert adapters) contain no user-identifying information. Removing the proxy artifact is both necessary and sufficient for complete user data removal from the inference system.
\end{invariant}

Importantly, this invariant is structural as opposed to statistical. While approximate unlearning methods provide probabilistic guarantees that user influence has been reduced below some threshold, Invariant~\ref{inv:separation} guarantees that user influence is \emph{architecturally absent} from shared components. In other words, the guarantee holds by construction as the system never permits user-specific gradients to flow into shared weights, so there is nothing to remove.

\subsection{Three-Layer Composition}
\label{sec:layers}

SEA combines three layers at inference time (Figure~\ref{fig:architecture}): a frozen base model that provides general capabilities, shared domain-expert LoRA adapters that provide specialized knowledge, and per-user proxy artifacts that provide deletable personalization.

\textbf{Base Layer.} The base layer is a frozen, quantized LLM that provides general language capabilities and is shared across all users. It contains no user-specific information by design, and the base weights are never modified during user interactions. Periodic retraining on aggregated data with differential privacy guarantees (DP-SGD \citep{abadi2016deep}) is a natural extension but is out of scope for this paper.

\textbf{Expert Layer.} A bank of $k$ domain-specific LoRA adapters $\mathcal{E} = \{E_1, \ldots, E_k\}$ provides specialized capabilities for distinct knowledge domains. Each expert $E_i = (B_i, A_i)$ is a low-rank adapter trained on curated domain corpora and shared across all users, with experts encoding domain knowledge only. At inference, experts combine via weighted linear combination (Equation~\ref{eq:expert-compose}):
\begin{equation}
  W_{\text{expert}} = W_{\text{base}} + \sum_{i=1}^{k} w_i \cdot B_i A_i
  \label{eq:expert-compose}
\end{equation}
where $\mathbf{w} \in \Delta^k$ (the probability simplex) are mixing coefficients determined per-query by a lightweight router.

\textbf{User Layer.} Each user $u$ has an isolated \emph{proxy artifact} $P_u$, which is a self-contained directory comprising three complementary personalization mechanisms, each stored as serialized tensors:

\begin{enumerate}
  \item \textbf{Routing bias vector} $\mathbf{b}_u \in \mathbb{R}^k$: A learned vector of domain affinity scores derived from user interaction patterns that shifts expert selection toward user-preferred domains. The bias is applied as a scaled additive adjustment with clamp-and-normalize:
  \begin{equation}
    \tilde{w}_i = w_{0,i} + \lambda\, b_{u,i}, \qquad
    w_i = \frac{\max(\tilde{w}_i,\, 0)}{\sum_j \max(\tilde{w}_j,\, 0)}
    \label{eq:bias}
  \end{equation}
  where $\mathbf{w}_0$ is the router's base distribution and $\lambda$ is a bias scale that prevents raw affinity values from overwhelming the base routing. If $\sum_j \max(\tilde{w}_j, 0) = 0$, the distribution falls back to uniform: $w_i = 1/k$.

  \item \textbf{Contrastive steering vectors} $\{s_u^\ell\}_{\ell \in \mathcal{L}}$ at a subset of intermediate layers $\mathcal{L}$: Computed via Contrastive Activation Addition \citealp{panickssery2024steering} from user preference pairs and injected additively into residual stream activations at inference:
  \begin{equation}
    \mathbf{h}^\ell \leftarrow \mathbf{h}^\ell + \gamma \, s_u^\ell
    \label{eq:steer}
  \end{equation}
  where $\gamma$ is a steering strength multiplier. These vectors encode stylistic preferences (verbosity, formality, technical depth) without modifying any model weights, making them particularly well-suited for deletable personalization.

  \item \textbf{Personal LoRA adapter} $L_u = (B_u, A_u)$: A low-rank adapter trained on user preference pairs. This adapter captures user-specific knowledge and response patterns that routing bias and steering alone cannot express, resulting in additional personalization. The rank is deliberately kept small to bound proxy size and maintain a clear separation guarantee. During personal LoRA training via DPO, the base model and expert adapter weights are then frozen, such that only the rank-4 personal LoRA parameters receive gradient updates, ensuring that user-specific gradients never flow into shared components.
\end{enumerate}

The proxy is operationally independent of shared weights at inference time, as it is a self-contained, deletable artefact whose removal then eliminates all user-specific influence from the system. However, note that the personal LoRA is conditioned on the shared model during DPO, where the base model serves as the reference, so the proxy's content reflects shared model state even though no user information flows in the reverse direction.

\subsection{Inference Pipeline}
\label{sec:pipeline}

Given query $q$ from user $u$, inference proceeds in five stages that combine the three layers into a single generation pass:

\begin{enumerate}
  \item \textbf{Route.} A lightweight router classifies $q$ into a domain distribution $\mathbf{w}_0 \in \Delta^k$ over the $k$ experts.

  \item \textbf{Bias.} The user's routing bias is applied via Equation~\ref{eq:bias}, shifting expert selection toward the user's preferred domains based on their accumulated interaction history.

  \item \textbf{Merge.} The weighted expert adapters and personal LoRA are combined into a single merged adapter applied to the base model.

  \item \textbf{Steer.} Forward hooks inject the user's steering vectors $\gamma \, s_u^\ell$ at layers $\ell \in \mathcal{L}$ via Equation~\ref{eq:steer}, modifying activations without changing any weights.

  \item \textbf{Generate.} Standard autoregressive decoding with the merged model produces the personalized output.
\end{enumerate}

\subsection{Deletion Protocol}
\label{sec:deletion}

SEA's deletion protocol exploits the architectural invariant (Invariant~\ref{inv:separation}) to reduce user removal to a simple filesystem operation with statistical verification. The key challenge we address is establishing that removing a user’s proxy artifact fully eliminates all user‑specific influence on model behavior.

To delete user $u$, the protocol proceeds in three steps:

\begin{enumerate}
  \item \textbf{Verify.} On held-out domain-generic prompts (not user-specific, to avoid circular verification): generate outputs in \emph{omission mode} (proxy not loaded) and compare token-frequency distributions against a cached non-personalized baseline (base model + experts, no proxy) via KL divergence. Verification uses a noise-calibrated threshold: the inter-sample KL divergence among unpersonalized generations provides an empirical noise floor $\hat{\sigma}_{\text{KL}}$ for stochastic decoding, and bypass is confirmed when
    \begin{equation}
      D_{\text{KL}}(p_{\text{unpers}} \| p_{\text{baseline}}) \;\leq\; \max\!\bigl(2\,\hat{\sigma}_{\text{KL}},\; \tau_{\min}\bigr)
      \label{eq:kl-verify}
    \end{equation}
    where $\tau_{\min} = 0.15$ nats is a hard floor that prevents unreasonably tight thresholds on low-variance queries. This makes verification self-calibrating: queries with high stochastic variance receive a proportionally wider acceptance band, eliminating false failures from sampling noise without weakening the guarantee for stable queries.

  \item \textbf{Delete.} Secure filesystem removal of the proxy directory $P_u$ (zero-overwrite).

  \item \textbf{Audit.} Log the deletion event, verification result, and timestamp for compliance trail.
\end{enumerate}

The architectural separation produces a direct payoff here. Without the proxy, the system's behavior is \emph{structurally equivalent in expectation} to the non-personalized baseline. The same code paths execute with the same weights, with the proxy simply not loaded. Verification exploits this architectural equivalence: omitting the proxy at inference time is functionally identical to deleting it, so the verify step confirms deletion behavior \emph{before} the irreversible delete step. The KL-divergence verification is therefore a sanity check confirming the architectural guarantee, not the privacy guarantee itself. The guarantee comes from the invariant: user information exists only in the proxy, and the proxy has been deleted. Cached baselines must be refreshed whenever shared components (base model or expert adapters) are updated; if a new base model is deployed, personal LoRA adapters must be regenerated.

\section{Experimental Setup}
\label{sec:setup}

We evaluate SEA across two base models, four domain experts, and four synthetic user profiles, targeting three evaluation dimensions: personalization quality, deletion completeness, and cross-user isolation. We first describe the experimental configuration and then present the results.

\textbf{Models.} We use two base models: Phi-3.5-mini-instruct (3.8B parameters) and Llama-3.1-8B-Instruct, both loaded in 4-bit NormalFloat (NF4) quantization via QLoRA \citep{dettmers2023qlora}. These models span a range of parameter counts to test whether the architectural properties hold across model scales.

\textbf{Expert Adapters.} Four domain experts ($k = 4$) are trained via supervised fine-tuning with TRL \citep{vonwerra2020trl}, all using rank 32, scaling factor $\alpha = 64$, applied to all attention projections (query, key, value, output): Security (Trendyol + OWASP-NVD, ${\sim}$76K examples), Code (CodeAlpaca + supplementary code instruction sets, capped at ${\sim}$50K examples), Data (synthetic text-to-SQL), and General (Alpaca, ${\sim}$52K examples). These experts are shared across all users and contain domain knowledge only.

\textbf{Synthetic User Profiles.} Four user profiles (\texttt{security\_expert}, \texttt{casual\_coder}, \texttt{data\_analyst}, \texttt{general\_user}) are each defined by domain affinity weights and positive/negative style traits. Proxy artifacts are generated through three mechanisms: routing bias via EMA from simulated interaction patterns ($\lambda = 0.5$), steering vectors via CAA from trait-aligned preference pairs at layers $\mathcal{L} = \{12, 16, 20\}$ with strength $\gamma = 1.0$, and personal LoRA (rank 4) via DPO \citep{rafailov2023direct} on preference pairs, using the base model as the DPO reference. The total proxy size is approximately 2--5~MB per user.

\textbf{Routing and Composition.} The expert router uses zero-shot entailment-based classification \citep{yin2019benchmarking} using BART-MNLI \citep{lewis2020bart} with keyword-based fallback (softmax temperature $T = 2.0$ for the fallback path). Adapter merging uses PEFT's \texttt{add\_weighted\_adapter} with \texttt{combination\_type="linear"} and a load-once lifecycle with deferred cleanup.

\textbf{Evaluation Protocol.} We conduct 70 evaluation runs per model (140 total) across 20 evaluation prompts (5 per domain).\footnote{Each evaluation run generates 7 bypass observations (a subset of query-user combinations selected from the held-out verification prompts). Phi-3.5-mini completed 68 runs (476 observations); Llama-3.1-8B completed 70 runs (490 observations). Two early Phi-3.5-mini runs were configuration tests that produced no bypass data.} Cached baselines ensure consistency across runs, and 95\% confidence intervals are reported via the $t$-distribution.

\textbf{Style trait match.} Style trait match is defined as the number of target style keywords detected in a personalized generation. Each user profile specifies a set of positive style traits as keywords (e.g., terms associated with verbosity, technical depth, or domain-specific vocabulary), and the metric counts how many appear in each output. The reported value is the mean count across all prompt-user-run observations (1,904 for Phi-3.5-mini, 1,960 for Llama-3.1-8B). The scale is profile-dependent: the security expert profile achieves a mean of 3.01 (Phi) and 1.02 (Llama), while the general user profile averages 0.21 and 0.28 respectively. Keyword presence is a necessary but not sufficient indicator of style alignment, as a response containing a target keyword may use it in a non-stylistic context. The metric should therefore be understood as a lower bound on non-match rather than a calibrated measure of style fidelity.

\section{Results}
\label{sec:results}

We organize results around three claims that jointly aim to validate the architectural design. First, we show that the proxy achieves measurable personalization (\S\ref{sec:personalization}), second, that the proxy removal restores baseline behavior (\S\ref{sec:separability}), and third that no cross-user leakage occurs (\S\ref{sec:isolation}). Together, these claims address the central question of whether architectural separation can simultaneously deliver personalization, deletability, and isolation.

\subsection{Personalization}
\label{sec:personalization}

The proxy measurably adapts model outputs without modifying shared weights. Table~\ref{tab:personalization} shows three distinct findings. First, routing bias successfully shifts expert selection toward each user's preferred domain (weight shift 0.052--0.088). Second, Jaccard similarity to the non-personalized baseline is low (0.236--0.316), indicating substantial output differentiation. Third, style trait matching is stronger for Phi-3.5-mini (1.71) than Llama-3.1-8B (0.63), an observed difference between these two specific models that should not be attributed to model size given $N{=}2$ and multiple confounds.

\begin{table}[h]
  \centering
  \caption{Personalization metrics across both base models. Weight shift measures the routing bias effect on expert selection. Jaccard similarity to baseline measures output overlap (lower = more personalized). Style trait match measures alignment with target user traits.}\label{tab:personalization}
  \small
  \begin{tabular}{lcc}
  \toprule
  Metric & Phi-3.5-mini & Llama-3.1-8B \\
  \midrule
  Weight shift & 0.052 $\pm$ 0.002 & 0.088 $\pm$ 0.003 \\
  Jaccard similarity & 0.236 $\pm$ 0.005 & 0.316 $\pm$ 0.005 \\
  Style trait match & 1.710 $\pm$ 0.101 & 0.629 $\pm$ 0.040 \\
  \bottomrule
  \end{tabular}
\end{table}

The three-mechanism proxy thus achieves moderate-to-strong personalization for Phi-3.5-mini and moderate personalization for Llama-3.1-8B, without touching shared weights. The personalization is present but deliberately moderate in scope, a consequence of the rank-4 constraint on the personal LoRA, which is the price of deletability and a central trade-off of our design. More expressive adapters would capture richer user preferences but would require more parameters, increasing proxy size and reducing the clarity of the separation guarantee. The security expert profile produces the strongest personalization signal (mean style trait match 3.01 on Phi-3.5-mini, with individual observations reaching 12), yet bypass verification for this profile's queries passes at rates comparable to lower-personalization profiles. The architecture does not trade deletion reliability for personalization intensity.

\subsection{Separability}
\label{sec:separability}

Next, we find that proxy removal restores baseline behavior, which confirms the architectural invariant. Table~\ref{tab:separability} shows two main results. First, mean KL divergence between unpersonalized and baseline outputs is approximately 0.21 nats for both models. Second, the 82--89\% noise-calibrated verification pass rate indicates that the vast majority of prompt-user combinations produce outputs statistically indistinguishable from the non-personalized baseline after proxy removal.

\begin{table}[h]
  \centering
  \caption{Deletion verification metrics. Verification pass rate is the fraction of prompt-user combinations where the unpersonalized-to-baseline KL divergence falls within the noise-calibrated threshold (Equation~\ref{eq:kl-verify}).}\label{tab:separability}
  \small
  \begin{tabular}{lcc}
  \toprule
  Metric & Phi-3.5-mini & Llama-3.1-8B \\
  \midrule
  Verified pass rate & 0.819 $\pm$ 0.035 & 0.892 $\pm$ 0.028 \\
  KL divergence & 0.217 $\pm$ 0.012 & 0.212 $\pm$ 0.006 \\
  \bottomrule
  \end{tabular}
\end{table}

\begin{figure*}[t]
  \centering
  \includegraphics[width=0.55\textwidth]{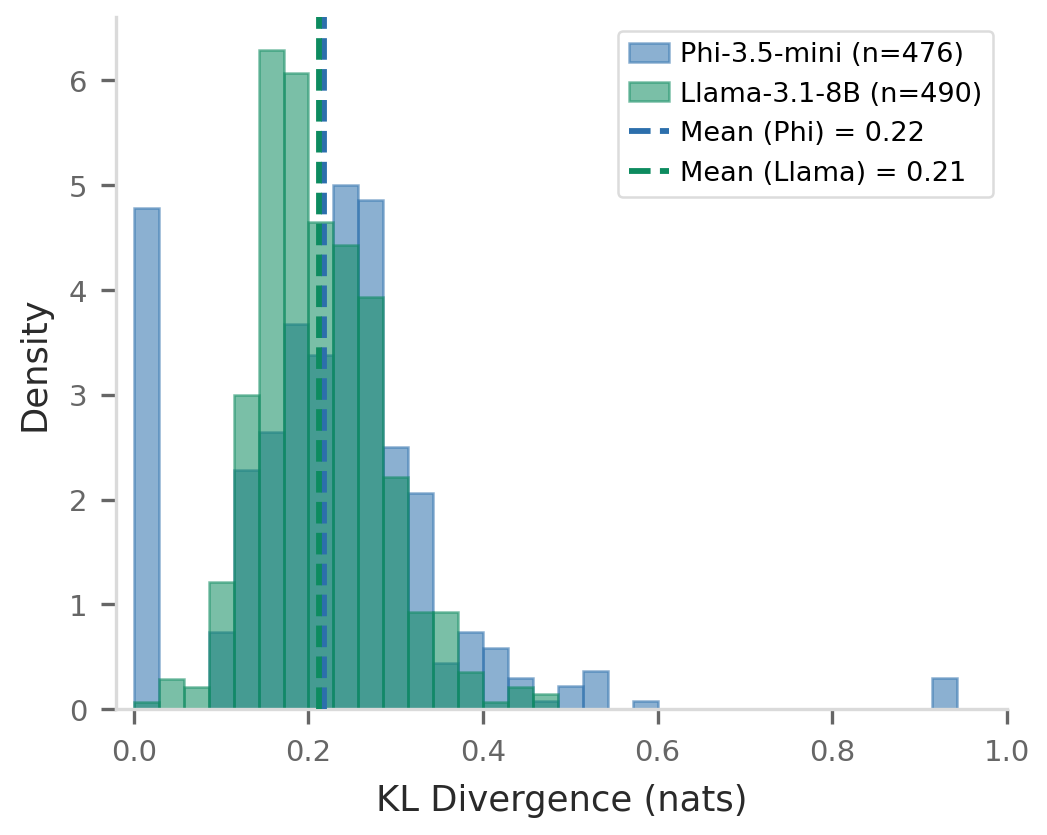}
  \caption{Distribution of unpersonalized-to-baseline KL-divergence scores across all prompt-user combinations for both base models (476 observations for Phi-3.5-mini, 490 for Llama-3.1-8B). Dashed lines mark the per-model mean. Verification uses a noise-calibrated per-query threshold (Equation~\ref{eq:kl-verify}) rather than a fixed cutoff, so no single threshold line is shown. The KL distribution is bimodal rather than gradual: verified observations cluster in [0.00, 0.30] and failures in [0.30, 0.94], with no ambiguous intermediate population. This sharp boundary is consistent with the structural guarantee, as proxy removal either fully eliminates user influence (the common case) or generation variance produces an outlier sample (the failure case), with no evidence of partial leakage.}
  \label{fig:kl-distribution}
\end{figure*}

Figure~\ref{fig:kl-distribution} shows the distribution of KL-divergence scores across all prompt-user combinations. Importantly, the deletion itself is deterministic and complete, as the proxy files are removed and the shared weights are untouched. The KL verification is a separate measurement that compares stochastic outputs from finite-length generations. By calibrating the acceptance threshold against the empirical inter-sample noise floor per query, the verification procedure accounts for the inherent variance of stochastic decoding: Queries that naturally produce high output variance receive a proportionally wider threshold, while stable queries are held to a tighter standard. The 11--18\% of cases that still exceed the noise-calibrated threshold likely reflect edge cases where generation variance is unusually high relative to the measured noise floor, not residual user influence in the weights.\footnote{A small number of Phi-3.5-mini observations produced degenerate (near-empty) outputs due to an inference configuration issue that did not affect Llama-3.1-8B runs. These observations yield artificially low KL values and are retained in the reported statistics for transparency. Filtering them would increase the mean KL slightly and marginally reduce the reported pass rate for Phi-3.5-mini.} The deletion verification thus provides empirical confirmation of the architectural guarantee, though the guarantee itself rests on the structural invariant rather than the verification metric.

\textbf{Threshold sensitivity.} The verification pass rate reported above depends on the $2\hat{\sigma}_{\text{KL}}$ multiplier in Equation~\ref{eq:kl-verify}. Table~\ref{tab:sensitivity} shows how the pass rate varies across multiplier settings. The hard floor $\tau_{\min}$ is inert across the tested range $[0.10, 0.25]$ because the empirical noise floor $\hat{\sigma}_{\text{KL}} \approx 0.15$ nats is stable across all query-user pairs (range $[0.146, 0.157]$), making the multiplier the sole active control. The floor would activate only if $\hat{\sigma}_{\text{KL}}$ dropped below $\tau_{\min}/\text{mult}$ (approximately 0.075 nats at the paper's $2\sigma$, $\tau_{\min} = 0.15$ configuration), which does not occur in this data. A single multiplier parameter therefore suffices for threshold calibration. This cross-query, cross-user, cross-model consistency was not guaranteed by the architecture and constitutes an empirical finding: the stochastic decoding noise floor is a property of the generation process, not of the personalization mechanism, which is what a structurally clean separation should produce.

\begin{table}[h]
  \centering
  \caption{Verification pass rate by $\sigma$ multiplier. The chosen $2\sigma$ configuration (bold) sits in the moderate region of a monotonic curve. Stricter deployments could tighten to $1.5\sigma$ at the cost of more false failures; those prioritizing operational stability could relax to $2.5\sigma$.}\label{tab:sensitivity}
  \small
  \begin{tabular}{lcc}
  \toprule
  Multiplier & Phi-3.5-mini ($n{=}476$) & Llama-3.1-8B ($n{=}490$) \\
  \midrule
  $1.0\sigma$ & 0.239 & 0.167 \\
  $1.5\sigma$ & 0.513 & 0.600 \\
  $\mathbf{2.0\sigma}$ & \textbf{0.819} & \textbf{0.892} \\
  $2.5\sigma$ & 0.929 & 0.984 \\
  $3.0\sigma$ & 0.971 & 0.994 \\
  \bottomrule
  \end{tabular}
\end{table}

Pass rates increase monotonically with no discontinuities. The deletion guarantee is independent of these parameters, as this analysis characterizes verification sensitivity as opposed to deletion completeness. The KL distributions across all observations have mean 0.218 (Phi) and 0.213 (Llama), with standard deviations of 0.132 and 0.070 respectively. Phi-3.5-mini has a heavier right tail (95th percentile 0.402 vs 0.340), which explains its lower pass rate at the same threshold.

\subsection{Isolation}
\label{sec:isolation}

Moreover, our results suggest that no cross-user leakage occurs between proxies. Table~\ref{tab:isolation} shows very low levels of contamination: 0.009 and 0.049 for Phi-3.5-mini and Llama-3.1-8B respectively, suggesting that one user's proxy does not influence another user's outputs. Cross-user output similarity is moderate (0.27--0.35) but expected, as users share the same base model and expert adapters. This similarity is structural and not leakage, reflecting the shared foundation rather than cross-user information flow.

\begin{table}[h]
  \centering
  \caption{Cross-user isolation metrics. Contamination measures excess inter-user similarity beyond the shared baseline.}\label{tab:isolation}
  \small
  \begin{tabular}{lcc}
  \toprule
  Metric & Phi-3.5-mini & Llama-3.1-8B \\
  \midrule
  Contamination & 0.009 $\pm$ 0.002 & 0.049 $\pm$ 0.005 \\
  Cross-user similarity & 0.271 $\pm$ 0.010 & 0.351 $\pm$ 0.007 \\
  \bottomrule
  \end{tabular}
\end{table}

Since proxies exist as isolated filesystem artifacts with no shared mutable state, this result follows from the architecture. However, we include it as empirical verification that the isolation invariant holds in practice under realistic generation conditions.

\textbf{Summary.} Taken together, the three claims are supported across both models with some between-model heterogeneity: Phi-3.5-mini shows stronger personalization and isolation, while Llama-3.1-8B shows stronger deletion verification rates. Llama-3.1-8B achieves a higher verification pass rate (89.2\% vs 81.9\%) with a substantially tighter KL distribution (std 0.070 vs 0.132), indicating that the deletion properties of the architecture do not degrade at the larger model scale. This shows that architectural separation achieves personalization with verified deletion and clean isolation, while the tradeoff between personalization expressiveness and deletability is explicit. The proxy's tunable parameters (personal LoRA rank, steering strength $\gamma$, routing bias scale $\lambda$) define a configuration space that could be explored to characterize this tradeoff, though the current evaluation uses a single configuration throughout.

\section{Discussion}
\label{sec:discussion}

\textbf{Contribution.} SEA sidesteps the machine unlearning problem rather than solving it. Machine unlearning is fundamentally hard because it attempts to undo an irreversible operation, the entanglement of user-specific gradients with shared weights. Even the most promising methods either require retraining or cannot guarantee complete removal. Architectural separation prevents entanglement in the first place, converting an intractable algorithmic problem into a tractable engineering one. The core tradeoff is explicit: A low-rank personal LoRA is less expressive than full fine-tuning, but the three-mechanism proxy compensates for this by providing complementary personalization channels (routing bias for domain preferences, steering vectors for stylistic preferences, and personal LoRA for residual patterns). The architecture's parameters (personal LoRA rank, steering strength $\gamma$, routing bias scale $\lambda$) define a per-deployment configuration space in which personalization fidelity can be traded against proxy size and separation clarity. Characterizing this tradeoff empirically, for instance by comparing rank-4 against rank-8 or rank-16 personal LoRA under the same deletion protocol, remains future work. A notable consequence of the separation invariant is that shared model components (the base model and expert adapters) can be released or audited without risk of user data exposure, since no user-specific information enters shared weights by construction. Moreover, it is important to note that our approach requires designing the system with deletion in mind from the start and cannot be retrofitted to existing models where user data has already been absorbed into weights.

\textbf{Findings.} Our evaluation across two base models shows three main results. First, the personal proxy produces measurable personalization, with users receiving responses that reflect their domain preferences and stylistic tendencies, with consistent shifts in routing weights and style trait alignment. Second, deletion verification works: When a user's proxy is removed, the system's outputs return to baseline behavior in 82--89\% of test cases, with the remaining failures attributable to normal generation randomness rather than lingering user influence (the architecture structurally guarantees that no trace of the user persists). Third, user isolation holds with one user's proxy not detectably influencing another user's outputs (contamination $\leq 0.05$ in point estimates). These results come with the inherent tradeoff that deletability limits how deeply the system can personalize, since user data must remain separable rather than being absorbed into shared model weights. We view this as a reasonable price for deployments where data deletion rights must be honored.

\textbf{Limitations and future work.} Several limitations constrain the current evaluation. The synthetic user profiles used here are placeholders for real-world preferences, and the four profiles are aligned to four distinct domains, representing the easiest possible configuration for isolation testing; overlapping-domain profiles (e.g., two security-focused users with different stylistic preferences) would provide a harder and more realistic test of cross-user isolation, though the structural separation guarantee is unaffected by profile design. The metrics (Jaccard similarity, keyword matching) capture basic textual overlap rather than subjective personalization quality as perceived by users in order to demonstrate the proof-of-concept. Second, the evaluation at 3.8--8B parameter scale is not intended to generalize to larger models, though the architectural invariant (separation of user data into a deletable proxy) holds by construction regardless of model size. Third, the current evaluation does not include an ablation study isolating the contribution of each proxy component (routing bias, steering vectors, personal LoRA individually); such an ablation would clarify which mechanisms drive personalization and deletion properties and is a natural next step. Additionally, while architectural separation eliminates the risk of user data being entangled in shared weights, the proxy artifact concentrates user behavioral information into a portable representation, creating an attack surface where an attacker need only exfiltrate a single directory rather than extract user influence from distributed weights. For open-source base models, including both models evaluated in this paper, an exfiltrated proxy could be loaded directly against a local copy. Non-transferability of exfiltrated proxies is therefore a hypothesis requiring empirical validation through cross-model transfer experiments, not a default assumption. Securing proxy artifacts through encryption at rest, access controls, and retention policies is necessary for end-to-end privacy and should be treated as a deployment requirement. Tractable deletion is also a dual-use capability, with the same mechanism that enables personal data removal also being easily applied to remove other content or proprietary knowledge from model integration, with implications for compliance auditing that merit careful analysis. Lastly, expert adapter training may not have converged, as loss plateaus were not reached during the experiments, suggesting that additional training could improve adapter quality.

The most immediate extension is applying DP-SGD to the gradient aggregation stage when updating shared expert adapters from user interaction data, which the architecture already supports by construction. Three practical constraints govern this extension: the computational overhead of per-sample gradient clipping, accelerated privacy budget exhaustion under sequential composition, and utility degradation in low-$\varepsilon$ regimes. Aggregating LoRA updates across a large user population prior to noise injection could provide privacy amplification, since individual contributions to the aggregate gradient would be attenuated by population scale. However, formal privacy amplification results depend on specific mathematical conditions, including Poisson subsampling of participants, bounded per-sample sensitivity, and particular composition theorems \citep{balle2018privacy,mironov2017renyi}, none of which have been verified for this architecture. Whether SEA's gradient aggregation satisfies these conditions, and whether the resulting $\varepsilon$-utility tradeoff is favorable in practice, are open empirical questions that require measuring privacy loss under varying $\varepsilon$ and population-size configurations through empirical attacks (model inversion, membership inference) against the updated shared model. Beyond DP-SGD, scaling to production multi-tenant workloads via adapter-serving frameworks such as S-LoRA and Punica, validating the privacy guarantees through longitudinal studies with real users and adversarial probes, and characterizing the tradeoff between personalization depth and proxy size are all natural next steps.

\bibliographystyle{unsrtnat}
\bibliography{references}

\end{document}